\documentclass[5p]{elsarticle}
\usepackage{algpseudocode}
\usepackage{algorithm}
\usepackage{amsmath}
\usepackage{amsfonts}
\usepackage{amssymb}

\usepackage[hidelinks]{hyperref}
\usepackage{float}

\journal{Journal of \LaTeX\ Templates}

\usepackage{framed,multirow}
\usepackage[export]{adjustbox}
\usepackage{subfig}
\usepackage{graphicx}
\usepackage{booktabs}

\begin{document}
\thispagestyle{empty}









\clearpage

\ifpreprint
  \setcounter{page}{1}
\else
  \setcounter{page}{1}
\fi
\begin{frontmatter}

\title{DocSegTr: An Instance-Level End-to-End Document Image Segmentation Transformer}

\fntext[myfootnote]{Authors contribute equally.}

\author[cvcuabaddress]{Sanket Biswas\fnref{myfootnote}\corref{mycorrespondingauthor}}

\cortext[mycorrespondingauthor]{Corresponding author}
\ead{sbiswas@cvc.uab.es}

\author[cvcuabaddress]{Ayan Banerjee\fnref{myfootnote}}\ead{abanerjee@cvc.uab.es}
\author[cvcuabaddress]{Josep Lladós}\ead{josep@cvc.uab.es}
\author[isiaddress]{Umapada Pal}
\ead{umapada@isical.ac.in}


\address[cvcuabaddress]{Computer Vision Center, Computer Science Department, Universitat Aut\`{o}noma de Barcelona, Spain.}
\address[isiaddress]{Computer Vision and Pattern Recognition Unit, Indian Statistical Institute, India}


\begin{abstract}
Understanding documents with rich layouts is an essential step towards information extraction. Business intelligence processes often require the extraction of useful semantic content from documents at a large scale for subsequent decision-making tasks. In this context, instance-level segmentation of different document objects (title, sections, figures etc.) has emerged as an interesting problem for the document analysis and understanding community. To advance the research in this direction, we present a transformer-based model called \emph{DocSegTr} for end-to-end instance segmentation of complex layouts in document images. The method adapts a twin attention module, for semantic reasoning, which helps to become highly computationally efficient compared with the state-of-the-art. To the best of our knowledge, this is the first work on transformer-based document segmentation. Extensive experimentation on competitive benchmarks like PubLayNet, PRIMA, Historical Japanese (HJ) and TableBank demonstrate that our model achieved comparable or better segmentation performance than the existing state-of-the-art approaches with the average precision of \textbf{89.4}, \textbf{40.3}, \textbf{83.4} and \textbf{93.3}. This simple and flexible framework could serve as a promising baseline for instance-level recognition tasks in document images.
\end{abstract}



\end{frontmatter}


\section{Introduction}
\label{s:intro}


Intelligent Document Processing (IDP) has grown in recent years owing to the advent of new digitization services in key industrial sectors like finance, insurance, law, banking, healthcare and so on. New Robotic Process Automation (RPA) tools in paperless offices have managed to automate the information extraction of document workflows, by building intelligent systems that can both read and understand. In recent times, state-of-the-art deep learning approaches have been leveraged for information extraction from both paper and digital documents. They achieve it by understanding spatial layout elements and text semantics as objects in document images, and the problem has been reformulated as Document Object Detection (DOD)\cite{Li_2020_CVPR}. But the task of document object detection is more focused on localizing layout elements of a document image such as tables, paragraphs, titles, figures etc. with bounding boxes and then predicting the semantic category for that particular object. For complex document layouts, extracting the layout structure with only bounding box information becomes quite difficult in case of overlapped object categories as highlighted in our recently published work \cite{biswas2021beyond}. Instance-level segmentation task provides a more precise way of extracting structured layouts, as it involves accurate prediction of masks (instances) at the pixel level. Modern successful approaches to solve the task are mostly based on CNN \cite{huang2019mask}. Although CNN's help to capture local features (low-level semantics) quite effectively from images, it performs sub-optimally in getting global semantic reasoning of features at the high-level. Also, these methods follow a top-down approach which is highly reliant on the object detector to detect regions with bounding boxes, and then predict per-pixel segmentation masks. This turns out to be a more computationally expensive strategy in terms of inference speed and also inferior segmentation performances in more complex structured layouts.

In this work, we hope to overcome the above drawbacks by proposing a new bottom-up approach to instance-level segmentation of document layouts, using the simple principle of dynamic instance mask generation inspired by \cite{wang2020solov2}, without any dependence on the bounding box locations and scales. This also leads to more efficiency in model inference time during the process of  the validation samples. Further, inspired by the power of transformers in document understanding \cite{appalaraju2021docformer,souibgui2022text} and natural language processing \cite{devlin2018bert} tasks in recent studies, we try to use them as global context aggregation module in our architecture pipeline. Transformers have the ability to capture dense long-range semantic dependencies with self-attention module, that aggregates both positional and feature information of the input domain. This is particularly beneficial in the detection of large overlapping objects as seen in this recent work \cite{carion2020end}. But they do not perform as well as CNNs do when extracting local features, which often leads to error-prone classification of smaller objects. In this work, we investigate the power of transformers as global context feature aggregator, for already obtained extracted features by passing the input document through a CNN backbone. Also, we take inspiration from Guo et. al. \cite{guo2021sotr} to implement an efficient twin-attention mechanism on our overall model pipeline, a sparse representation of the conventional self-attention matrix in Vaswani et al. \cite{vaswani2017attention}. 

The overall contributions of our work can be summarized into two folds: 1) A simple Document image Segmentation Transformer (\emph{DocSegTr}) framework is introduced for an end-to-end bottom-up instance-level segmentation baseline that works effectively to analyze and segment complex structured document layouts that leverages the effectiveness of both CNNs and transformers in a single-stage architectural pipeline, without depending on detection of instances with bounding boxes. 2) \emph{DocSegTr} adapts the twin-attention based transformer module from Guo et. al. \citep{guo2021sotr} for semantic reasoning, which helps it to become the most computationally efficient for inference on document images compared with the existing state-of-the-art approaches based on Mask-RCNN \citep{biswas2021beyond}. 

The rest of this paper is organized as follows. In Section \ref{s:sota} we review the state of the art. The \emph{DocSegTr} model  is described in Section \ref{s:method}. Section \ref{s:experiments} provides the experimental evaluation, including the ablation studies. Section \ref{s:results} contains a discussion of the extensive experimentation that has been conducted. Finally, in Section \ref{s:conclusion} we draw the conclusions and propose future research lines.


\section{Related Work}
\label{s:sota}
Information  extraction  from   digital  documents require the spatial understanding and relational reasoning between different layout components of a page such as tables, text blocks (paragraphs), figure regions, title heads and so on. Several pieces of  work have approached the problem of page object localization, ranging from classical heuristic rule-based document layout segmentation approaches to the recent deep learning-based  Document Object Detection (DOD). In this section, we make an effort to overview the state-of-the-art according to different methodological schemes.


Document layout segmentation using heuristic methods is classified mainly under three different categories: top-down, bottom-up and hybrid strategies. The bottom-up method \citep{ asi2015simplifying} used pixels as basic components, and performed operations like merging and grouping to form a larger homogeneous region. On the other hand, top-down method \citep{ journet2005text} relied on splitting the whole document image iteratively into different regions, until a definite standard column or block was obtained. Although bottom-up methods were more dexterous and could be applied to documents having far more diverse layouts, it was computationally expensive. Top-down approaches could instead provide more efficiency in computation, but only limited to processing specific types, like documents with a manhattan-based layout. Hence, hybrid method \citep{tran2015hybrid} were adopted in literature which could combine both top-down and bottom-up cues to generate better results. Prior to deep learning era,  rule-based segmentation technique \citep{fang2011table} were also used profoundly to solve the table detection problem. 


With the advent of the deep learning revolution, Convolutional Neural Networks (CNNs) became the primary component of the state-of-the-art techniques for document layout segmentation. The goal was mostly to have a strong object detection model for documents to segment layout elements as objects following the strong baselines of Faster-RCNN \citep{ren2015faster}, Mask-RCNN \citep{he2017mask} and Retinanet \citep{lin2017focal} for detecting natural scene objects. DeepDeSRT  \citep{schreiber2017deepdesrt} became the first popular object detection approach for table detection and structure recognition in documents (both scanned and born-digital images). A novel image transformation strategy was applied on the documents before feeding them to the Faster-RCNN model. Later, fully convolutional neural networks (FCNNs) \citep{he2017multi} have been used for detecting multiple object categories (tables and figures) on pages. Oliviera et. al. \citep{oliveira2018dhsegment} used a similar FCNN-based framework for pixel-wise segmentation on historical documents. Similarly, Biswas et al. \cite{biswas2021beyond} perform instance level document segmentation on scientific reports as well as historical documents with the mask-RCNN baseline beside the object detection task. This architecture contains a MLP head after a strong CNN baseline for object detection at bounding box level and an FCN head for pixel-wise segmentation. To unify and combine the above deep learning-based layout segmentation baselines for all competitive DOD benchmarks, LayoutParser \citep{shen2021layoutparser} toolkit was introduced. Recently, a new cross-domain Document Object Detection (DOD) benchmark was also established in \citep{Li_2020_CVPR} to apply domain adaptation strategies for DOD to solve the domain shift problem. 
   


Natural language models in deep learning era started gaining more prominence with the self-attention and positional embedding mechanisms addressed in Vaswani et. al. \citep{vaswani2017attention} leading to the origin of transformers. Later, BERT \citep{devlin2018bert} language model has been used in multi-modal state-of-the-art Visual Document Understanding (VDU) tasks like named entity recognition and key-value pair information extraction in scanned receipts and invoices. LayoutLM \citep{xu2020layoutlm} used joint learning of text, layout and visual features of document images with BERT-like model baseline, to obtain state-of-the-art results in VDU related tasks. Other recent  approaches \citep{kim2021donut, li2021selfdoc} in VDU using transformers include joint pre-training on text, layouts and image regions of a page to solve VDU tasks like form understanding, receipt understanding and document visual question answering\citep{mathew2021docvqa}.  

Motivated by the recent breakthrough of transformers in document analysis and understanding systems, we propose the first end-to-end document image segmentation transformer, that achieves superior performance on standard instance-level segmentation benchmark datasets. A fair comparative experimental study has also been provided in this work with our previous baseline approach \citep{biswas2021beyond} and LayoutParser \citep{shen2021layoutparser} which are the standard benchmark approaches till date.

\section{Method}
\label{s:method}
\begin{figure}
  \includegraphics[width=\columnwidth]{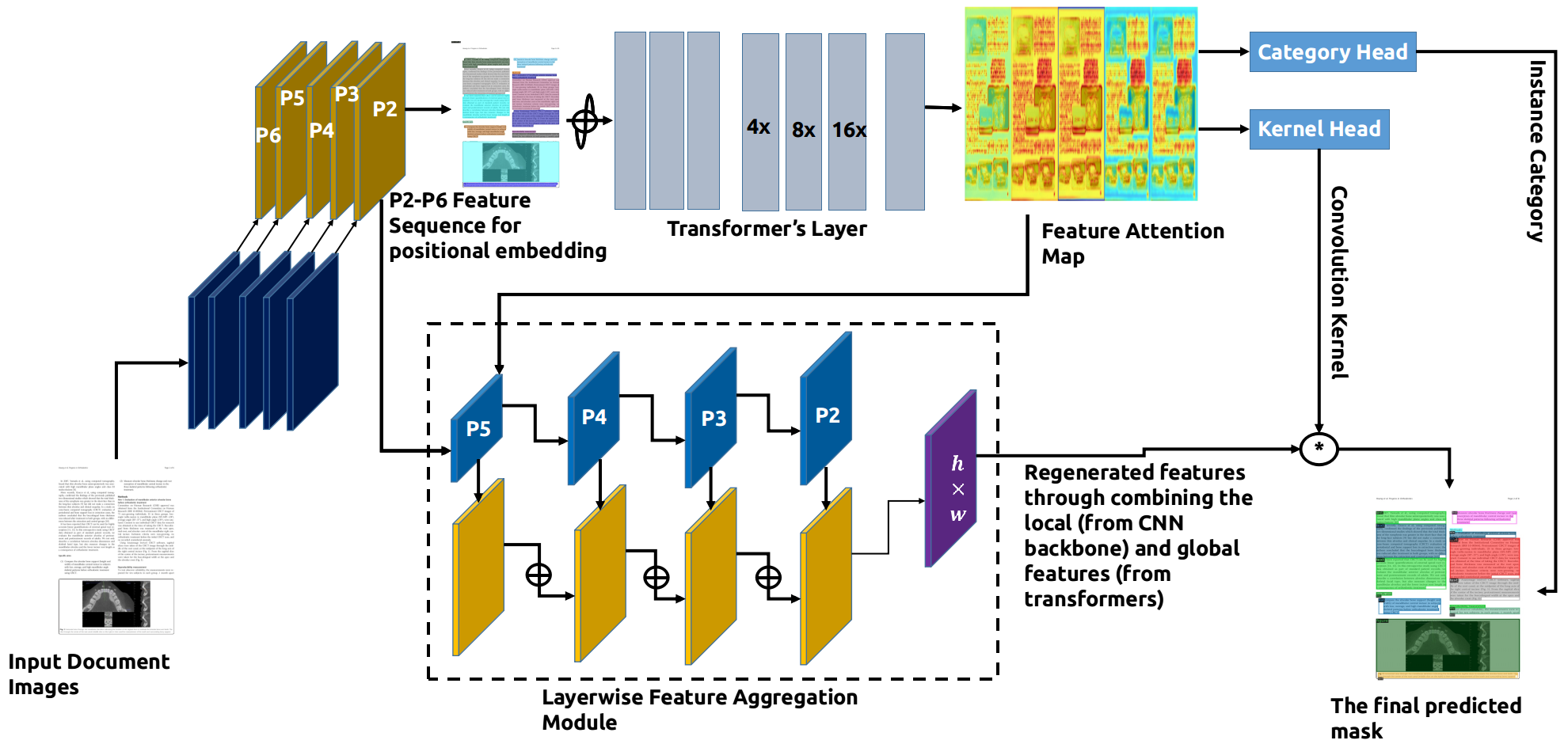}
\caption{\textbf{DocSegTr} builds on a simple CNN feature extractor with FPN on the input document image. The multi-scaled feature maps (P2-P6) from FPN are combined with positional embedding information to feed into transformer layers, to predict document instances and generate corresponding kernel dynamically. The LFAM combines the local FPN features and global transformer feature from P5 to segment the instances on the document image.}
\label{fig: method}       
\end{figure}

In order to accomplish the instance-level document layout segmentation task, a hybrid CNN-based transformer pipeline called \emph{DocSegTr} has been proposed which can simultaneously learn to capture variable layout elements (text, table, figures, titles etc.) and also long-range dependencies between them. It is a direct end-to-end segmentation paradigm, that divides input feature maps into patches, and then predicts per-patch document category. \emph{DocSegTr} essentially comprises three different modules: 1) a backbone of CNN with Feature Pyramid Networks (FPNs) for feature extraction, particularly at lower-level (i.e. local features), from the input document image. 2) a transformer to capture contextual semantic reasoning on top of those feature elements, which is connected with functional heads to predict semantic class category (class head) and their corresponding convolutional kernel as output to the kernel head (this operation is done after flattening the multi-scale features as well as adding positional embeddings) and 3) a Layerwise Feature Aggregation module (LFAM) to compound and aggregate the feature representations obtained from local convolutional and global transformer backbones to generate the final feature map. Then a dynamic convolution activity is performed between the generated feature map and the corresponding convolutional kernel to obtain the predicted segmented instances in the document image. The general structure of our architectural pipeline is portrayed as shown in Fig. \ref{fig: method}.

\subsection{Document Segmentation Transformer (\emph{DocSegTr})}
The proposed \emph{DocSegTr} consists of three modules: a self attention mechanism to tackle the long-range dependencies, Transformer layer, and a Functional heads dedicated to subsequent mask prediction for instance segmentation. Each module is further described in the following subsections.

\subsubsection{Attention Mechanism:}

The self-attention module is a key part of transformer models, which intrinsically catches full-document image context and learns important correlations between different feature components in the extracted feature information from the document image. But the self-attention module implemented in the original transformer architecture proposed in \cite{vaswani2017attention} is computationally quite expensive both in terms of memory and time.
To alleviate this issue, we propose the twin attention system as introduced in \cite{guo2021sotr} to decompose the original attention matrix into a sparse one. In this strategy, attention is first computed for each column independently. This helps to aggregate contextual reasoning on a horizontal scale. 
Then the same process is repeated for the vertical scale, by computing similar attention for each row. Then the resultant row and column attentions (for vertical and horizontal scales respectively) are connected sequentially to a global attention module, to cover necessary feature information along two different dimensions (global and local).  

In the instance-level segmentation task, given the feature maps $f_i \in \mathbb{R}^{h \times w \times c}$ at each layer $i$ of FPN, \emph{DocSegTr} first divides the feature maps into $n \times n$ patches $p_i \in \mathbb{R}^{n \times n \times c}$ and later stacks them into fixed components for vertical and level bearings. Positional 
embeddings are supplemented with the above patch components to hold positional information, implying that the position embeddings for row and column matrices are adopted in the self-attention mechanism. To make sure that both modules (row and column attention matrices) work with multi-head attention and connect across multiple layers, all the sub-layers in the twin attention considers to produce the same $n \times n \times c$ output. This approach can successfully reduce the memory and time complexity from standard $\Theta (h \times w)^2$ to $\Theta (h \times w^2+w \times h^2)$. The memory and computational complexity here is expressed in a more general form of H and W instead of N since the twin attention can process the input of any resolution, not limited to the square tensor.

\subsubsection{Transformer Layer:}

In this section, we present three distinctive transformer layers in view of the encoder as our fundamental structure block (as outlined in model pipeline Fig. \ref{fig: method}). The first transformer layer resembles the original transformer encoder utilized in \citep{vaswani2017attention}, which considers two different components: 1) a multi-head attention unit after a layer norm \citep{ba2016layer}, 2) a multi-layer perceptron (MLP) block after layer norm.  Moreover, a residual association \citep{he2016deep} is utilized to connect the two components. Finally, a multi-dimensional feature sequence is obtained as the result of the $K$ sequential associations of the transformer layers, to make predictions in the subsequent functional heads described in the next subsection.

In the \emph{DocSegTr}, the original transformer layer has been adapted from \citep{vaswani2017attention} and the multi-head attention has been substituted with the twin attention mechanism proposed by Guo et. al \citep{guo2021sotr}.  The key intuition behind this approach is that twin-attention helps to  model global feature dependencies(on two different scales), and also in a sparse manner that saves computation.   
\subsubsection{Functional Heads}
The resultant sequence of feature maps obtained from the transformer layers are then fed to the two functional heads adapted in \emph{DocSegTr}. The category head includes a multi-layered perceptron (MLP) classifier layer to produce a $n \times n \times q_c$ result, where $q_c$ is the quantity of classes. It produces the semantic category prediction of a document instance (title, figure, list etc.).


The kernel head is also composed of a linear layer, in correspondence with the category head to yield a $n \times n \times b$ tensor for resulting veil age, where the tensor means the $n \times n$ convolutional kernels with $b$ number of parameters. This generated kernel is eventually responsible for predicting masks dynamically over the input document image as shown in Fig.\ref{fig: method} architectural pipeline. The focal loss \citep{lin2017focal} has been used to classify the predicted document instances. 

\subsection{Instance-level Segmentation with Mask features}
Here, we highlight how the instance-level segmentation strategy is carried out for the proposed \emph{DocSegTr} model. The first submodule highlights the feature generation strategy for instance masks combining CNN and transformer (both local and global features) units. The second submodule describes how the final mask prediction is carried out to segment the input document layouts.  
\subsubsection{Mask Feature Generation} To construct mask feature representations for instance-level segmentation of document layouts, a straightforward way is to make predictions on each feature map at different scales \citep{ju2021adaptive}. However, it increases time and memory complexities. To overcome this issue, a Layerwise Feature Aggregator Module (LFAM) has been adapted as shown in Fig. \ref{fig: method} to combine the multi-scaled features at every FPN level and transformer to a unified mask feature. The relatively low-resolution feature maps with positional information from the P5 block of transformer module is  fused with feature maps obtained from P2-P4 in FPN to execute a fusion, that encodes both global(from transformers) and local(from FPN) information from the document image at multiple scales of feature hierarchy. After the processed P2-P5 are concatenated together, a point-wise convolution and upsampling technique is executed to generate the final unified mask $h \times w$ feature maps.
\subsubsection{Instance Mask Prediction:}Finally for instance mask prediction on the document, \emph{DocSegTr} generates the mask for every patch by performing dynamic convolution operations on the above-unified feature maps. This dynamic prediction strategy has been inspired from \citep{wang2020solov2}. Given predicted convolution kernels $k \in \mathbb{R}^{n \times n \times b}$ from the kernel head, each kernel is responsible for the mask generation of the instance in the corresponding patch. The detailed operation can be expressed in equation (1).
\begin{equation}
    M_f^{h \times w \times n \times n} = f^{h \times w \times c} * k^{n \times  n \times b}
\end{equation}
Where * demonstrates the convolution activity, $M_f$ is the final generated mask with a component of $h \times w \times n \times n$. It
is worth mentioning that the value of $b$ relies upon the shape of the kernel, in other words, b equivalents $\theta^2*c$, where $\theta$ is the size of the kernel. The final instance segmentation mask is generated using the Matrix NMS \cite{huang2020nms} and regularisation is controlled by the Dice Loss \cite{wang2020improved}.
\begin{figure*}[htp]
\captionsetup[subfloat]{farskip=0pt,captionskip=0pt}
\subfloat[PublayNet]{%
  \includegraphics[width=0.5\columnwidth,height=4cm]{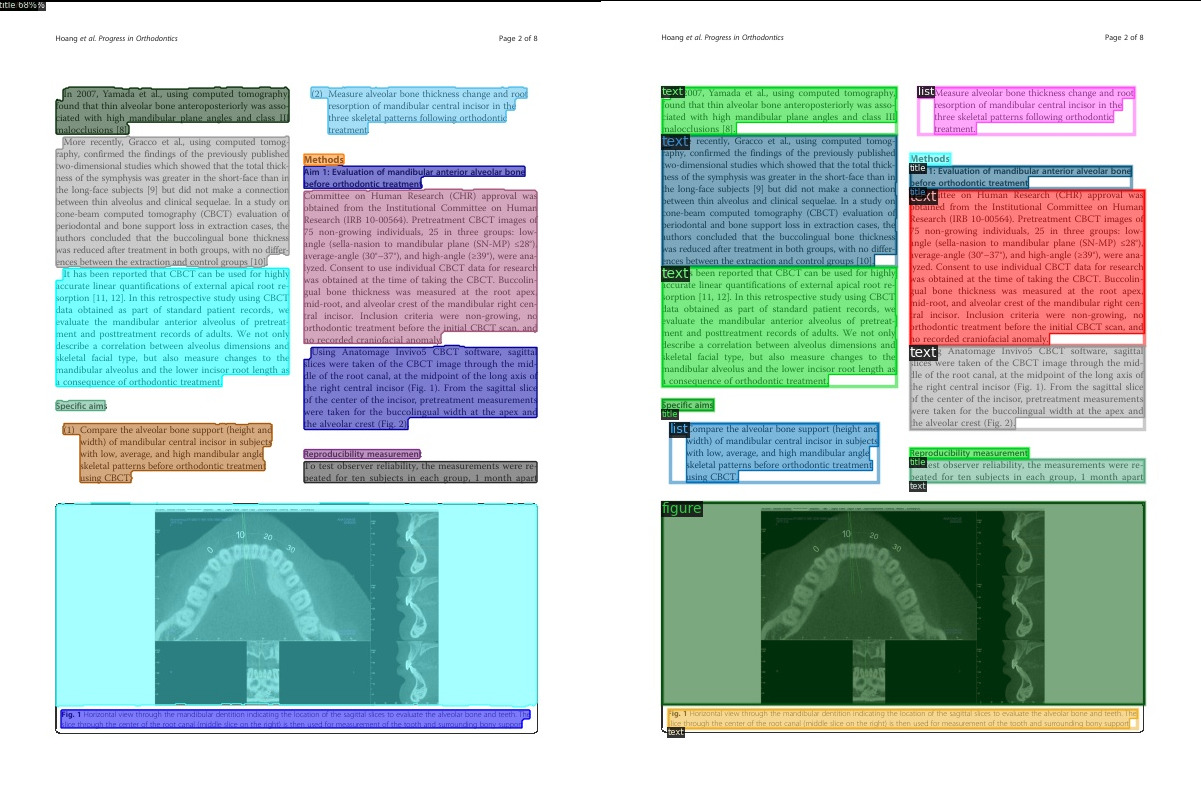}%
}
\subfloat[PRIMA]{%
  \includegraphics[width=0.5\columnwidth,height=4cm]{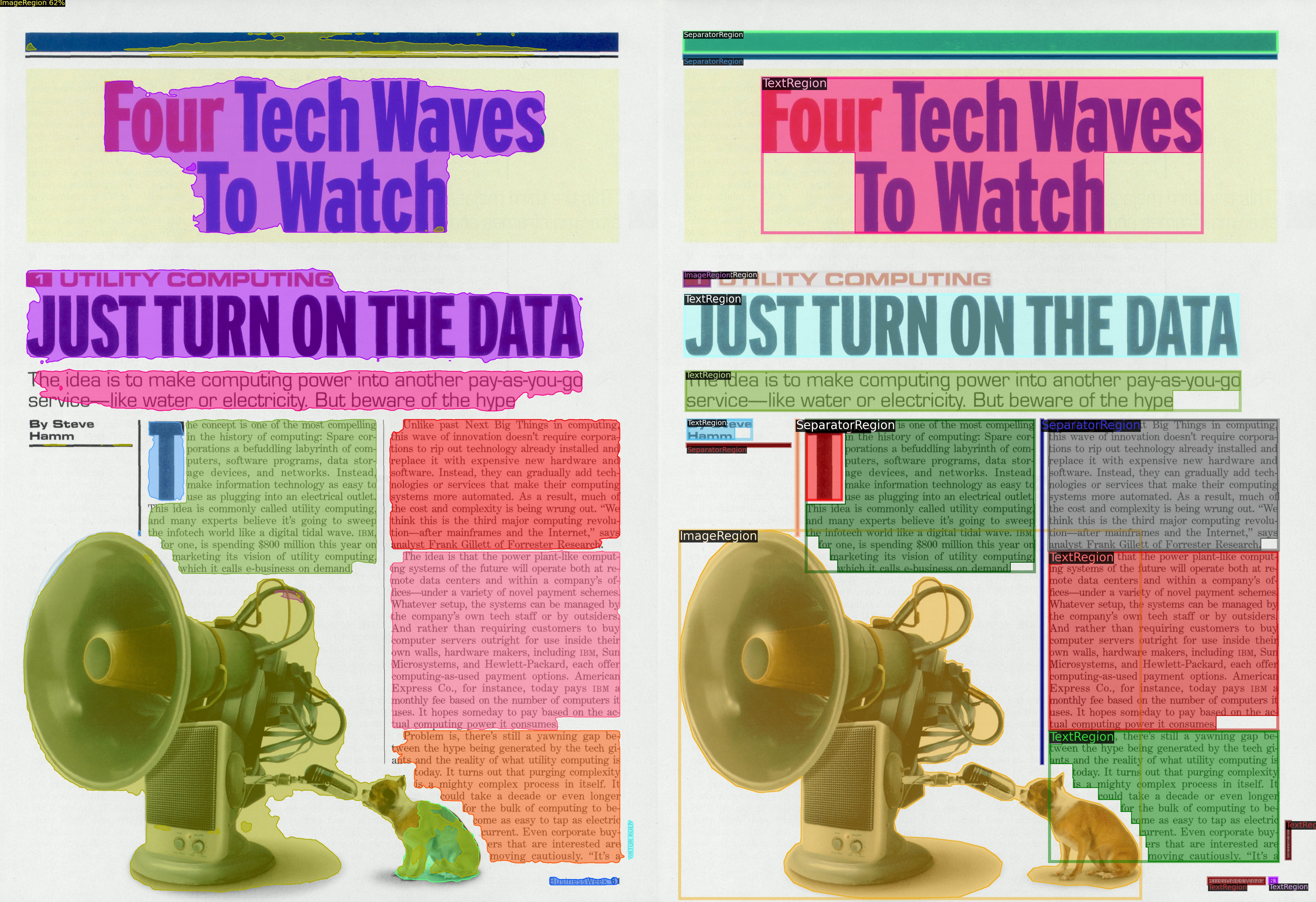}%
}
\subfloat[Historical Japanese (HJ)]{%
  \includegraphics[width=0.5\columnwidth,height=4cm]{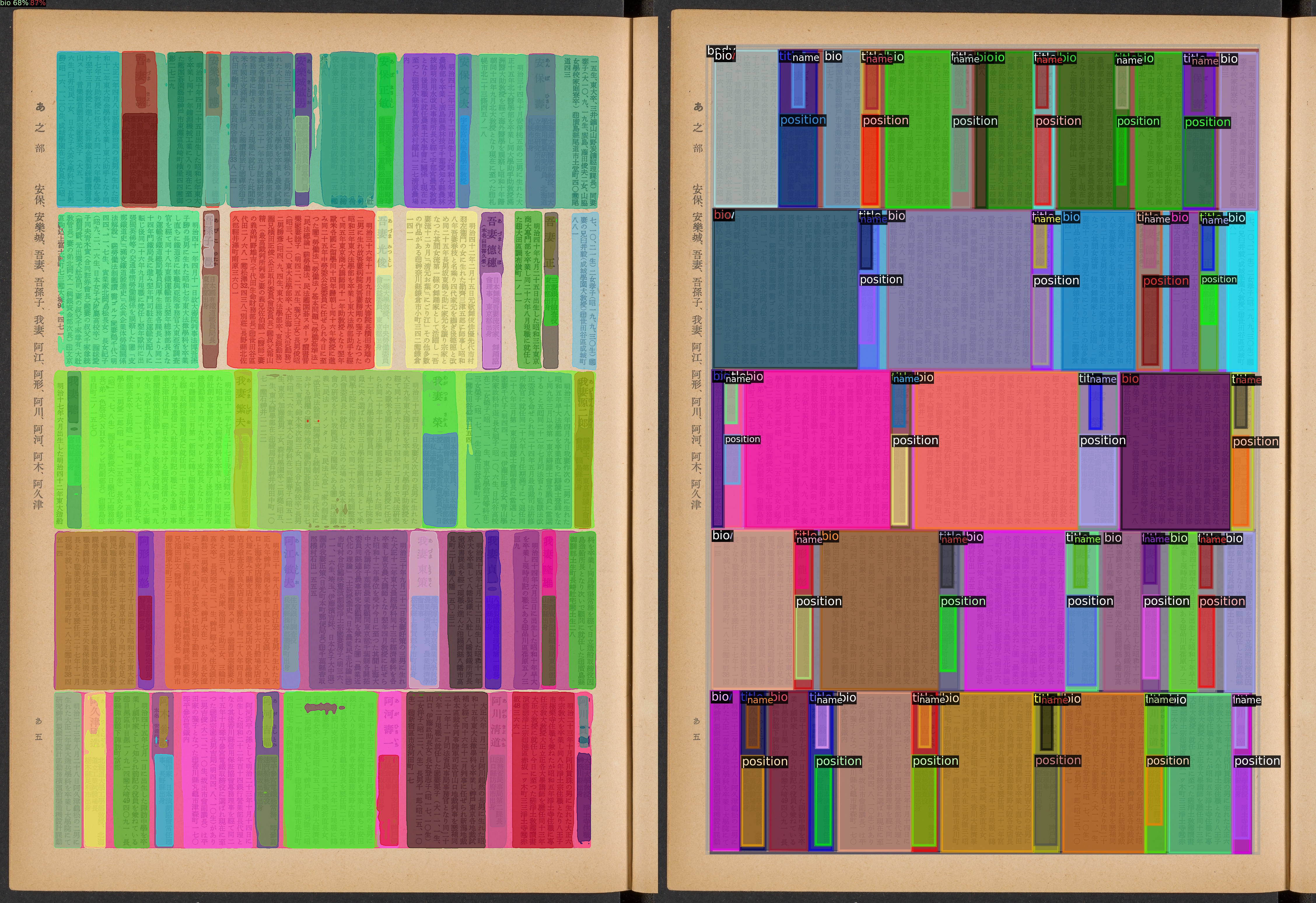}
}
\subfloat[TableBank]{%
  \includegraphics[width=0.5\columnwidth,height=4cm]{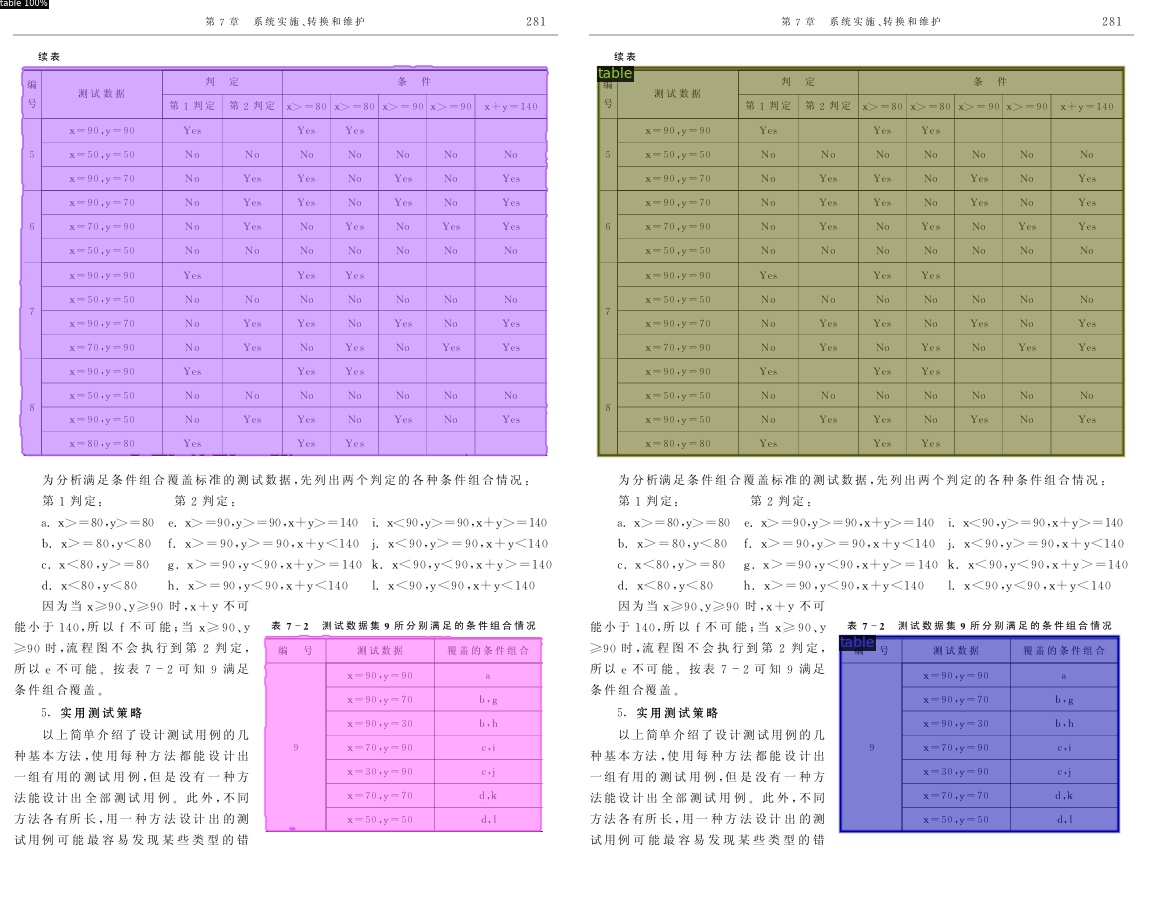}
}
\caption{Qualitative analysis of the DocSegTr framework on the four applied benchmarks (\textbf{Left:} Predicted layout \textbf{Right:} Ground-truth)}
\label{fig:examples_detection_all}
\end{figure*}

\section{Experiments}
\label{s:experiments}
For the purpose of validation, we have considered some important benchmark datasets (PublayNet, PRIMA, HJ and TableBank) with different document object categorization. Our experimental evaluation suggests that the proposed DocSegTr method advances the state-of-the-art. Moreover, extensive ablation studies have been performed to show the contribution of some important elements of the architecture. All the code and benchmarking models will be publicly available at:
\url{https://github.com/biswassanket/DocSegTr}.



\subsection{Dataset Description \& Evaluation Measures}

Lack of standard public benchmarks has always been a concern for the DLA community. With the advent of document understanding principles and the transformation of the problem motivation over the years, it lead to the release of large-scale annotated datasets like PubLayNet \citep{zhong2019publaynet}, HJ \citep{shen2020large}, TableBank \citep{prasad2020cascadetabnet} as well as small scale PRIMA \citep{clausner2019icdar2019}  which we use for evaluating our proposed segmentation approach in this work (Please refer to the Table 1 for detailed description).   

\begin{table}[ht]
\caption{Experimental dataset description (instance level)}
\centering
\resizebox{\columnwidth}{!}{
\begin{tabular}{@{}cccccccccccc@{}}
\toprule
\multicolumn{3}{c}{\textbf{PublayNet}}           & \multicolumn{3}{c}{\textbf{PRIMA}}               & \multicolumn{3}{c}{\textbf{Historical Japanese}} & \multicolumn{3}{c}{\textbf{TableBank}}           \\ \midrule
\textbf{Object} & \textbf{Train} & \textbf{Eval} & \textbf{Object} & \textbf{Train} & \textbf{Eval} & \textbf{Object} & \textbf{Train} & \textbf{Eval} & \textbf{Object} & \textbf{Train} & \textbf{Eval} \\ \midrule
Text            & 2,343,356      & 88,625        & Text            & 6401           & 1531          & Body            & 1443           & 308           & Table           & 2835           & 1418          \\ \midrule
Title           & 627,125        & 18,801        & Image           & 761            & 163           & Row             & 7742           & 1538          & -               & -              & -             \\ \midrule
Lists           & 80,759         & 4239          & Table           & 37             & 10            & Title           & 33,637         & 7271          & -               & -              & -             \\ \midrule
Figures         & 109,292        & 4327          & Math            & 35             & 7             & Bio             & 38,034         & 8207          & -               & -              & -             \\ \midrule
Tables          & 102,514        & 4769          & Separator       & 748            & 155           & Name            & 66,515         & 7257          & -               & -              & -             \\ \midrule
-               & -              & -             & other           & 86             & 25            & Position        & 33,576         & 7256          & -               & -              & -             \\ \midrule
-               & -              & -             & -               & -              & -             & Other           & 103            & 29            & -               & -              & -             \\ \midrule
\textbf{Total}  & 3,263,046      & 120,761       & \textbf{Total}  & 8068           & 1891          & \textbf{Total}  & 181,097        & 31,866        & \textbf{Total}  & 2835           & 1418          \\ \bottomrule
\end{tabular}
}
\end{table}

The most generic way to evaluate the correctness of the predicted instance (document category) for instance-level segmentation task is the Intersection over Union (IoU) score. The mean Average Precision (mAP) score for the overall model is computed by finding the mean of APs at different IoU thresholds (0.5 to 0.95 with a step size of 0.05), as used in standard Microsoft COCO \citep{lin2014microsoft} benchmark evaluation for instance segmentation. Also, model performance for evaluating each categorical document instance has also been computed as proposed in \citep{biswas2021beyond}.


\subsection{Ablation Study}

Extensive ablation studies were conducted in the context of instance-level segmentation to quantify the significance of every component of our overall model framework and to justify its usage for segmenting different layout elements. In this study, we take a look at the different feature extraction CNN backbones as we compare DocSegTr's performance with ResNeXt-101 compared to ResNet-101. Table \ref{tab:ablation_1_pub} clearly justifies the usage of ResNeXt-101 for our task. Not only that, we conducted a study with different CNN feature extraction backbones with and without Deformable Convolutional Networks (DCNs). Results in Table  \ref{tab:ablation_1_pub} highlight the significance of DCNs as an important component for the model backbone. The high performance gain is attributed to the fact that DCN's help to gain more richer local information from the document images.

But the interesting behaviour of the DocSegTr has been observed with the transformer layers and its attention mechanism. If we only use the ResNet-FPN backbone with DCN layers it only extracts the visual features and no contextual reasoning has been performed. That's why the performance of the setup is very poor (~5\% AP). In the next step, we have utilized the transformer layers without the self-attention mechanism (i.e. a MLP block after layer normalization) which shows a significant improvement of the performance. However, still the architecture doesn't provide its best results which depicts the importance of the self-attention mechanism for end-to-end instance layer segmentation.

\begin{figure}[htp]
\centering
\captionsetup[subfloat]{farskip=0pt,captionskip=0pt}
\subfloat[ResNet-FPN]{%
  \includegraphics[width=\columnwidth]{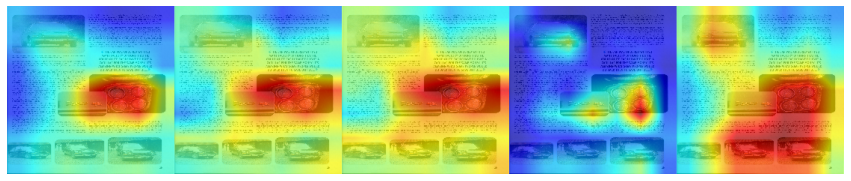}%
}

\subfloat[ResNet-FPN+DCN]{%
  \includegraphics[width=\columnwidth]{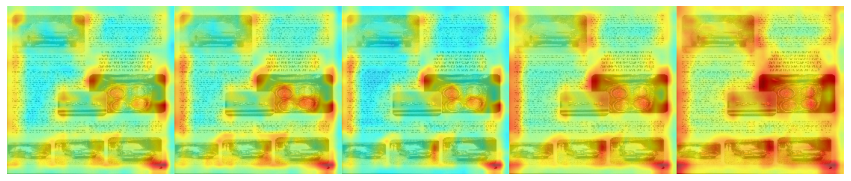}%
}

\subfloat[ResNet-FPN+DCN+Transformer]{%
  \includegraphics[width=\columnwidth]{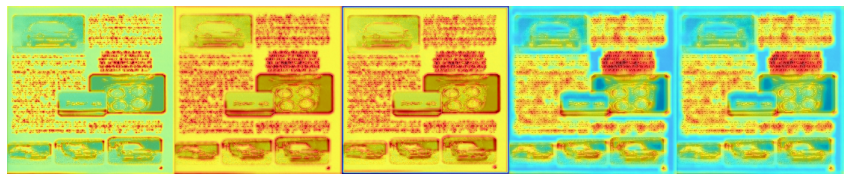}
}

\caption{Illustration of \textbf{attention maps} generated in three different case studies which demonstrates the utility of using self-attention mechanism for DocSegTr}
\label{fig:attention_map}
\end{figure}

The attention map obtained in Fig. \ref{fig:attention_map} depicts that the ResNet-FPN backbone can only extract the visual features of the large objects and skip smaller objects. When the DCN layers has been introduced it mainly focus in the edge reconstruction and unable to make significant improvement over the previous backbone. However, with the Transformer it focus the small as well as large objects and with sharp edges which describes the relevance of each individual step.

All the ablation experiments have been performed on the PRIMA dataset as it is proven that the transformers performed significantly well for larger datasets as the they need a large amount of training data points. This particular study has been conducted on a much smaller dataset to find how transformers adjust its training parameters in such situation. That is why we run an inference of the weights pre-trained with PublayNet dataset on the PRIMA benchmark without any finetuning and get an AP of ~15\% which depicts the robustness of the DocSegTr.


\begin{table}[ht]
\centering%
\setlength{\tabcolsep}{6pt}
\caption{Ablation Study of DocSegTr}
\captionsetup{skip=0pt}
\resizebox{\columnwidth}{!}{
\begin{tabular}{@{}cccc@{}}
\toprule
\textbf{Model backbone}                                                                                & \textbf{AP} & \textbf{AP@0.5} & \textbf{AP@0.75} \\ \midrule
\multicolumn{4}{c}{\textbf{ResNet vs ResNeXt}}                                                                                                            \\ \midrule
ResNet-101-FPN                                                                                         & 20.12       & 31.32           & 16.78            \\ \midrule
ResNeXt-101-FPN                                                                                        & 32.59       & 58.62           & 29.73            \\ \midrule
\multicolumn{4}{c}{\textbf{Deformable Convolution Networks}}                                                                                              \\ \midrule
ResNeXt-101-FPN                                                                                        & 32.59       & 58.62           & 29.73            \\ \midrule
ResNeXt-101-FPN(+DCN)                                                                                  & 33.21       & 49.13           & 27.39            \\ \midrule
\multicolumn{4}{c}{\textbf{Importance of Transformer (Contextual Reasoning)}}                                                                             \\ \midrule
without transformer                                                                                    & 5.21        & 7.12            & 3.22             \\ \midrule
\begin{tabular}[c]{@{}c@{}}without self-attention heads\\  (but using transformer layers)\end{tabular} & 29.14       & 41.23           & 20.22            \\ \midrule
DocSegTr                                                                                               & 40.31       & 59.72           & 29.54            \\ \bottomrule
\end{tabular}
}
\label{tab:ablation_1_pub}
\end{table}

\subsection{Implementation details}
We train DocSegTr with SGD setting the underlying learning pace of 0.001 with a steady warm-up of 1000 cycles and utilizing weight rot of $10^-5$ and Nesterov momentum of 0.9. For our removal tests, we train for 300K iterations with a learning rate drop by an element of 10 at 210K and 250K, individually. Except if indicated, all models are prepared on 2 Nvidia A40 GPUs of 48G RAM (take around 4-5 days), with batch size 8. The Python language is utilized for programming and the profound learning systems utilized are PyTorch and Detectron2.

\section{Results and Discussion}\label{s:results}
In this section, we have tried to provide some deep insights into the experimental results we have achieved both qualitatively and quantitatively. We shall discuss them in the following lines according to the datasets we have used to evaluate our proposed method.
\subsection{Qualitative Analysis}\label{ss:qualitative}

The segmentation results obtained by DocSegTr have been displayed in Fig. \ref{fig:examples_detection_all}. As shown in the example test cases, DocSegTr is able to segment instances of different layout elements (tables, figures, text regions etc.) quite effectively on the provided four different case studies of validation set samples. Looking at the first example document in Fig. \ref{fig:examples_detection_all}(a), the predicted instance masks look quite sharp and precise as we see the layouts are mostly regular and double-column depicted in ground truth. Also, for overlapped caption and figure in top region of the page, DocSegTr seems to perform correctly to segment the instances. In the second example Fig. \ref{fig:examples_detection_all}(b) the layout is more challenging compared to the first one as it is the page of a magazine. We can observe that the predicted instances are correct but not as sharp as the previous one and also miss one separator region, because the objects get smaller in ground truth example and not enough data to train as well as suffering with class imbalance problem. The third test case \ref{fig:examples_detection_all}(c) is the most challenging one of all as the layout of the page of historical documents, which is multi-column and the objects are small with very less spacing between the layout elements. However, the proposed model can predict accurate segmentation as compared to the ground truth mask elements. This dataset also has a very few samples but doesn't suffer with class imbalance problem. As a result, the bigger objects (even with some portion of overlapping) seems to be well-segmented along with smaller regions. Lastly, a simple downstream task of table detection has been demonstrated from a TableBank example shown in Fig. \ref{fig:examples_detection_all}(d). As tables are large objects and comparatively easier to segment due to their non-overlapping nature, DocSegTr manages to segment layouts with much higher performance on this dataset than the rest.  The following section \ref{ss:quantitative} justifies the quantitative reasons behind the performance of DocSegTr.

\subsection{Quantitative Analysis}\label{ss:quantitative}
As shown in Table \ref{tab:publaynet_final_results} the final performance of the DocSegTr model in terms of standard mean Average Precision(mAP) is computed overall and also categorically for every individual document object category (tables, list, text, title, figure). The results are quite interesting as they justify the qualitative results studied in the previous subsection. The APs of larger regions like lists, table and figure perform better than the previous Mask-RCNN based baseline models proposed in \cite{biswas2021beyond}. But the smaller regions have sub par APs when compared with the other one. This is attributed to the fact that DocSegTr has scope for improving performance specially segmenting those kind of layouts regions. Overall, DocSegTr achieves the state-of-the-art as seen in the Table \ref{tab:publaynet_final_results} compared with our previous work.

\begin{table*}[ht]
\caption{Quantitative performance evaluation of DocSegTr on the four different benchmarks}

\centering
\setlength{\tabcolsep}{6pt}
\resizebox{\textwidth}{!}{%
\begin{tabular}{@{}cccccccccccccccc@{}}
\toprule
\multicolumn{4}{c}{\textbf{PublayNet}}                                                                                                         & \multicolumn{4}{c}{\textbf{PRIMA}}                                                                                                               & \multicolumn{4}{c}{\textbf{Historical Japanese}}                                                    & \multicolumn{4}{c}{\textbf{TableBank}}                                                                                                         \\ \midrule
Object  & \begin{tabular}[c]{@{}c@{}}Layout\\  Parser\end{tabular} & \begin{tabular}[c]{@{}c@{}}Biswas \\  et al. \cite{biswas2021beyond}\end{tabular} & DocSegTr & Object    & \begin{tabular}[c]{@{}c@{}}Layout\\  Parser\end{tabular} & \begin{tabular}[c]{@{}c@{}}Biswas \\ et al.  \cite{biswas2021beyond}\end{tabular} & DocSegTr & Object   & LayoutParser & \begin{tabular}[c]{@{}c@{}}Biswas \\  et al. \cite{biswas2021beyond}\end{tabular} & DocSegTr & Object  & \begin{tabular}[c]{@{}c@{}}Layout\\  Parser \end{tabular} & \begin{tabular}[c]{@{}c@{}}Biswas \\  et al. \cite{biswas2021beyond}\end{tabular} & DocSegTr \\ \midrule
Text    & 90.1                                                                                                                             & 90.6                                                                                                                                                                       & 89.9     & Text      & 83.1                                                                                                                             & 77.2                                                                                                                                                                       & 70.8     & Body     & 99.0         & 99.6                                                                                                                                                                       & 99.0     & Table   & 91.2                                                                                                                             & 91.7                                                                                                                                                                       & 93.3     \\ \hline
Title   & 78.7                                                                                                                             & 81.8                                                                                                                                                                       & 73.6     & Image     & 73.6                                                                                                                             & 68.1                                                                                                                                                                       & 50.1     & Row      & 98.8         & 99.6                                                                                                                                                                       & 99.1     &         &                                                                                                                                  &                                                                                                                                                                            &          \\ \hline
Lists   & 75.7                                                                                                                             & 82.1                                                                                                                                                                       & 89.5     & Table     & 95.4                                                                                                                             & 82.4                                                                                                                                                                       & 42.5     & Title    & 87.6         & 91.3                                                                                                                                                                       & 93.2     &         &                                                                                                                                  &                                                                                                                                                                            &          \\ \hline
Figures & 95.9                                                                                                                             & 97.1                                                                                                                                                                       & 97.5     & Math      & 75.6                                                                                                                             & 55.6                                                                                                                                                                       & 26.5     & Bio      & 94.5         & 94.4                                                                                                                                                                       & 94.7     &         &                                                                                                                                  &                                                                                                                                                                            &          \\ \hline
Tables  & 92.8                                                                                                                             & 95.1                                                                                                                                                                       & 96.6     & Separator & 20.6                                                                                                                             & 17.2                                                                                                                                                                       & 9.6      & Name     & 65.9         & 68.1                                                                                                                                                                       & 70.3     &         &                                                                                                                                  &                                                                                                                                                                            &          \\ \hline
        &                                                                                                                                  &                                                                                                                                                                            &          & other     & 39.7                                                                                                                             & 22.8                                                                                                                                                                       & 17.4     & Position & 84.1         & 86.2                                                                                                                                                                       & 87.4     &         &                                                                                                                                  &                                                                                                                                                                            &          \\ \hline
        &                                                                                                                                  &                                                                                                                                                                            &          &           &                                                                                                                                  &                                                                                                                                                                            &          & Other    & 44.0         & 34.8                                                                                                                                                                       & 43.7     &         &                                                                                                                                  &                                                                                                                                                                            &          \\ \hline
AP      & 86.7                                                                                                                             & 89.3                                                                                                                                                                       & 89.4     & AP        & 64.7                                                                                                                             & 56.2                                                                                                                                                                       & 40.3     & AP       & 81.6         & 82.0                                                                                                                                                                       & 83.1     & AP      & 91.2                                                                                                                             & 91.7                                                                                                                                                                       & 93.3     \\ \hline
AP@0.5  & 97.2                                                                                                                             & 97.7                                                                                                                                                                       & 97.8     & AP@0.5    & 77.6                                                                                                                             & 67.3                                                                                                                                                                       & 59.7     & AP@0.5   & -            & 89.0                                                                                                                                                                       & 90.1     & AP@0.5  & 94.2                                                                                                                             & 94.9                                                                                                                                                                       & 98.5     \\ \hline
AP@0.75 & 93.8 & 95.3                                                                                                                                                                       & 95.4     & AP@0.75   & 71.6                                                                                                                             & 61.9                                                                                                                                                                       & 33.5     & AP@0.75  & -            & 87.8                                                                                                                                                                       & 88.1     & AP@0.75 & 92.1                                                                                                                             & 92.8                                                                                                                                                                       & 94.9\\ \bottomrule
\end{tabular}
}
\label{tab:publaynet_final_results}
\end{table*}

\section{Conclusion}\label{s:conclusion}
In this work, we proposed the Document Image Segmentation Transformer (DocSegTr) for instance-level segmentation of complex layout elements. To our knowledge, this bottom-up transformer-based baseline is the first work to explore instance-level document segmentation in a box-free manner. Further, we showcase a comprehensive study justifying the suitability of DocSegTr in different domains of layout samples. In the tested settings, DocSegTr demonstrates superior performance when compared with conventional approaches, specially in segmentation performance for larger document objects in an image. There is a large future scope towards this direction as transformers can prove to be highly beneficial for overlapped layout objects, but they do not show much improvement for smaller regions. Our work provides a simple and flexible baseline, that promises for further improvement. 

\bibliography{main}

\begin{thebibliography}{10}
\expandafter\ifx\csname url\endcsname\relax
  \def\url#1{\texttt{#1}}\fi
\expandafter\ifx\csname urlprefix\endcsname\relax\def\urlprefix{URL }\fi
\expandafter\ifx\csname href\endcsname\relax
  \def\href#1#2{#2} \def\path#1{#1}\fi

\bibitem{Li_2020_CVPR}
K.~Li, C.~Wigington, C.~Tensmeyer, H.~Zhao, N.~Barmpalios, V.~I. Morariu,
  V.~Manjunatha, T.~Sun, Y.~Fu, Cross-domain document object detection:
  Benchmark suite and method, in: Proceedings of the IEEE Conference on CVPR,
  2020.

\bibitem{biswas2021beyond}
S.~Biswas, P.~Riba, J.~Llad{\'o}s, U.~Pal, Beyond document object detection:
  instance-level segmentation of complex layouts, IJDAR, 2021.

\bibitem{huang2019mask}
Z.~Huang, L.~Huang, Y.~Gong, C.~Huang, X.~Wang, Mask scoring r-cnn, in:
  Proceedings of the IEEE Conference on CVPR, 2019, pp. 6409--6418.

\bibitem{wang2020solov2}
X.~Wang, R.~Zhang, T.~Kong, L.~Li, C.~Shen, Solov2: Dynamic and fast instance
  segmentation, NIPS, 2020.

\bibitem{appalaraju2021docformer}
S.~Appalaraju, B.~Jasani, B.~U. Kota, Y.~Xie, R.~Manmatha, Docformer:
  End-to-end transformer for document understanding, ICCV, 2021.

\bibitem{souibgui2022text}
M.~A. Souibgui, S.~Biswas, A.~Mafla, A.~F. Biten, A.~Forn{\'e}s, Y.~Kessentini,
  J.~Llad{\'o}s, L.~Gomez, D.~Karatzas, Text-diae: Degradation invariant
  autoencoders for text recognition and document enhancement, arXiv preprint
  arXiv:2203.04814.

\bibitem{devlin2018bert}
J.~Devlin, M.-W. Chang, K.~Lee, K.~Toutanova, Bert: Pre-training of deep
  bidirectional transformers for language understanding, arXiv preprint
  arXiv:1810.04805.

\bibitem{carion2020end}
N.~Carion, F.~Massa, G.~Synnaeve, N.~Usunier, A.~Kirillov, S.~Zagoruyko,
  End-to-end object detection with transformers, in: ECCV, 2020.

\bibitem{guo2021sotr}
R.~Guo, D.~Niu, L.~Qu, Z.~Li, Sotr: Segmenting objects with transformers, in:
  ICCV, 2021.

\bibitem{vaswani2017attention}
A.~Vaswani, N.~Shazeer, N.~Parmar, J.~Uszkoreit, L.~Jones, A.~N. Gomez,
  {\L}.~Kaiser, I.~Polosukhin, Attention is all you need, in: NIPS, 2017.

\bibitem{asi2015simplifying}
A.~Asi, R.~Cohen, K.~Kedem, J.~El-Sana, Simplifying the reading of historical
  manuscripts, in: Proceedings of the ICDAR, 2015.

\bibitem{journet2005text}
N.~Journet, V.~Eglin, J.-Y. Ramel, R.~Mullot, Text/graphic labelling of ancient
  printed documents, in: Proceedings of the ICDAR, 2005, pp. 1010--1014.

\bibitem{tran2015hybrid}
T.~A. Tran, I.-S. Na, S.-H. Kim, Hybrid page segmentation using multilevel
  homogeneity structure, in: Proceedings of the 9th International Conference on
  Ubiquitous Information Management and Communication, 2015, pp. 1--6.

\bibitem{fang2011table}
J.~Fang, L.~Gao, K.~Bai, R.~Qiu, X.~Tao, Z.~Tang, A table detection method for
  multipage pdf documents via visual seperators and tabular structures, in:
  ICDAR, 2011.

\bibitem{ren2015faster}
S.~Ren, K.~He, R.~Girshick, J.~Sun, Faster r-cnn: Towards real-time object
  detection with region proposal networks, in: NIPS, 2015.

\bibitem{he2017mask}
K.~He, G.~Gkioxari, P.~Doll{\'a}r, R.~Girshick, Mask r-cnn, in: Proceedings of
  the ICCV, 2017, pp. 2961--2969.

\bibitem{lin2017focal}
T.-Y. Lin, P.~Goyal, R.~Girshick, K.~He, P.~Doll{\'a}r, Focal loss for dense
  object detection, in: CVPR, 2017.

\bibitem{schreiber2017deepdesrt}
S.~Schreiber, S.~Agne, I.~Wolf, A.~Dengel, S.~Ahmed, Deepdesrt: Deep learning
  for detection and structure recognition of tables in document images, in:
  ICDAR, 2017.

\bibitem{he2017multi}
D.~He, S.~Cohen, B.~Price, D.~Kifer, C.~L. Giles, Multi-scale multi-task fcn
  for semantic page segmentation and table detection, in: Proceedings of the
  ICDAR, Vol.~1, 2017, pp. 254--261.

\bibitem{oliveira2018dhsegment}
S.~A. Oliveira, B.~Seguin, F.~Kaplan, dhsegment: A generic deep-learning
  approach for document segmentation, in: ICFHR, 2018.

\bibitem{shen2021layoutparser}
Z.~Shen, R.~Zhang, M.~Dell, B.~C.~G. Lee, J.~Carlson, W.~Li, Layoutparser: A
  unified toolkit for deep learning based document image analysis, in: ICDAR,
  2021, pp. 131--146.

\bibitem{xu2020layoutlm}
Y.~Xu, M.~Li, L.~Cui, S.~Huang, F.~Wei, M.~Zhou, Layoutlm: Pre-training of text
  and layout for document image understanding, in: ACM SIGKDD, 2020.

\bibitem{kim2021donut}
G.~Kim, T.~Hong, M.~Yim, J.~Park, J.~Yim, W.~Hwang, S.~Yun, D.~Han, S.~Park,
  Donut: Document understanding transformer without ocr, arXiv preprint
  arXiv:2111.15664.

\bibitem{li2021selfdoc}
P.~Li, J.~Gu, J.~Kuen, V.~I. Morariu, H.~Zhao, R.~Jain, V.~Manjunatha, H.~Liu,
  Selfdoc: Self-supervised document representation learning, in: CVPR, 2021,
  pp. 5652--5660.

\bibitem{mathew2021docvqa}
M.~Mathew, D.~Karatzas, C.~Jawahar, Docvqa: A dataset for vqa on document
  images, in: WACV, 2021.

\bibitem{ba2016layer}
J.~L. Ba, J.~R. Kiros, G.~E. Hinton, Layer normalization, arXiv preprint
  arXiv:1607.06450.

\bibitem{he2016deep}
K.~He, X.~Zhang, S.~Ren, J.~Sun, Deep residual learning for image recognition,
  in: Proceedings of the IEEE Conference on CVPR, 2016, pp. 770--778.

\bibitem{ju2021adaptive}
M.~Ju, J.~Luo, Z.~Wang, H.~Luo, Adaptive feature fusion with attention
  mechanism for multi-scale target detection, Neural Computing and
  Applications, 2021.

\bibitem{huang2020nms}
X.~Huang, Z.~Ge, Z.~Jie, O.~Yoshie, Nms by representative region: Towards
  crowded pedestrian detection by proposal pairing, in: CVPR, 2020.

\bibitem{wang2020improved}
L.~Wang, C.~Wang, Z.~Sun, S.~Chen, An improved dice loss for pneumothorax
  segmentation by mining the information of negative areas, IEEE Access, 2020.

\bibitem{zhong2019publaynet}
X.~Zhong, J.~Tang, A.~J. Yepes, Publaynet: largest dataset ever for document
  layout analysis, in: Proceedings of the ICDAR, 2019, pp. 1015--1022.

\bibitem{shen2020large}
Z.~Shen, K.~Zhang, M.~Dell, A large dataset of historical japanese documents
  with complex layouts, in: Proceedings of the IEEE Conference on CVPRW, 2020,
  pp. 548--549.

\bibitem{prasad2020cascadetabnet}
D.~Prasad, A.~Gadpal, K.~Kapadni, M.~Visave, K.~Sultanpure, Cascadetabnet: An
  approach for end to end table detection and structure recognition from
  image-based documents, in: CVPRW, 2020, pp. 572--573.

\bibitem{clausner2019icdar2019}
C.~Clausner, A.~Antonacopoulos, S.~Pletschacher, Icdar2019 competition on
  recognition of documents with complex layouts-rdcl2019, in: Proceedings of
  the ICDAR, 2019, pp. 1521--1526.

\bibitem{lin2014microsoft}
T.-Y. Lin, M.~Maire, S.~Belongie, J.~Hays, P.~Perona, D.~Ramanan,
  P.~Doll{\'a}r, C.~L. Zitnick, Microsoft coco: Common objects in context, in:
  Proceedings of the ECCV, 2014.

\end{thebibliography}
\end{document}